\documentclass{IEEEtran4PSCC}

\ifCLASSINFOpdf
   \usepackage[pdftex]{graphicx}

\else

   \usepackage[dvips]{graphicx}

\fi

\usepackage[cmex10]{amsmath}
\usepackage{multirow}
\usepackage{cite}
\usepackage{multicol}
\usepackage[htt]{hyphenat}  
\usepackage{amsmath}
\usepackage{titlesec}

\titlespacing*{\subsection}{0pt}{0.3em}{0.2em}

\allowdisplaybreaks[2]
\makeatletter
\let\old@ps@headings\ps@headings
\let\old@ps@IEEEtitlepagestyle\ps@IEEEtitlepagestyle
\def\psccfooter#1{%
    \def\ps@headings{%
        \old@ps@headings%
        \def\@oddfoot{\strut\hfill#1\hfill\strut}%
        \def\@evenfoot{\strut\hfill#1\hfill\strut}%
    }%
    \def\ps@IEEEtitlepagestyle{%
        \old@ps@IEEEtitlepagestyle%
        \def\@oddfoot{\strut\hfill#1\hfill\strut}%
        \def\@evenfoot{\strut\hfill#1\hfill\strut}%
    }%
    \ps@headings%
}
\makeatother

\psccfooter{%
        \parbox{\textwidth}{\hrulefill \\ \small{24th Power Systems Computation Conference} \hfill \begin{minipage}{0.2\textwidth}\centering \vspace*{4pt} \includegraphics[scale=0.06]{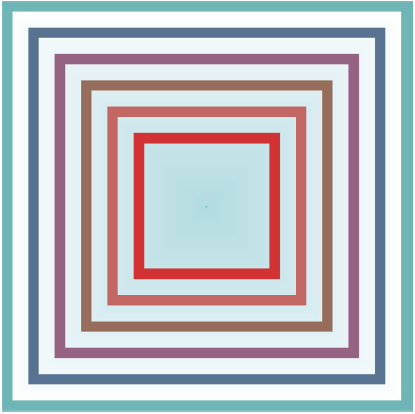}\\\small{PSCC 2026} \end{minipage} \hfill \small{Limassol, Cyprus --- June 8 -- June 12, 2026}}%
}

\begin{document}

\title{Residual Correction Models for AC Optimal Power Flow Using DC Optimal Power Flow Solutions}
 \author{\IEEEauthorblockN{
 Muhy Eddin Za'ter\IEEEauthorrefmark{1}, 
 Bri-Mathias Hodge\IEEEauthorrefmark{1} and Kyri Baker\IEEEauthorrefmark{1}} 
 \IEEEauthorblockA{\IEEEauthorrefmark{1} Electrical, Computer \& Energy Engineering and Sustainable Energy Institute\\ University of Colorado Boulder}
 }

\maketitle

\begin{abstract}
Solving the nonlinear AC optimal power flow (AC OPF) problem remains a major computational bottleneck for real-time grid operations. In this paper, we propose a residual learning paradigm that uses fast DC optimal power flow (DC OPF) solutions as a baseline, and learns only the nonlinear corrections required to provide the full AC-OPF solution. The method utilizes a topology-aware Graph Neural Network with local attention and two-level DC feature integration, trained using a physics-informed loss that enforces AC power-flow feasibility and operational limits. Evaluations on OPFData for 57-, 118-, and 2000-bus systems show around 25\% lower MSE, up to 3× reduction in feasibility error, and up to 13X runtime speedup compared to conventional AC OPF solvers. The model maintains accuracy under N-1 contingencies and scales efficiently to large networks. These results demonstrate that residual learning is a practical and scalable bridge between linear approximations and AC-feasible OPF, enabling near real-time operational decision making.
\end{abstract}


\section{Introduction and Background}
\label{sec:introduction}

The field of power systems optimization has been a central and long standing challenge since the early development of electricity markets and operations \cite{zhu2009optimization}. Historically, this field started with solving the Economic Dispatch (ED) problem, where system operators attempted to determine the allocation of generation resources to minimize operating costs while meeting demand \cite{wood2013power}. However, as transmission networks expanded in size and complexity, ED alone proved insufficient because it ignored physical network constraints such as line losses, voltage limits, and stability \cite{cain2012history}. Events such as the 1965 Northeast blackout in the United States \cite{vassell1990northeast} underscored the importance of incorporating grid physical constraints into operational planning, motivating the integration of security constraints into dispatch through contingency analysis and operating limits, which evolved into Security-Constrained Economic Dispatch (SCED) \cite{loehr2017good,vargas2002tutorial}. Yet, these early approaches were based on simplified models that could not capture the full complexity of power flow such as reactive power behavior and voltage stability \cite{wood2013power}. This limitation led to the formulation of the Optimal Power Flow (OPF) problem, which embeds AC network equations and operational constraints \cite{aravena2023recent}. Although AC-OPF offers the most accurate representation of the grid, its nonlinear and nonconvex nature makes it computationally expensive and often intractable for real time applications \cite{nair2022computational}.

To address this computational burden, approximations and relaxations have been developed. The most widely used is DC-OPF, which approximates the AC equations by neglecting reactive power, voltage magnitudes changes, and line losses, sacrificing accuracy for tractability \cite{frank2012optimal}. More advanced approaches include adaptive linearizations and convex relaxations such as second-order cone programming, aiming to balance solution quality and computational speed \cite{cain2012history,castillo2013computational}. While effective for many operational contexts, these methods struggle to guarantee AC feasibility, leaving a trade-off between computational efficiency and accuracy \cite{babiker2025optimal}.

These approximations were generally sufficient for traditional grids dominated by dispatchable synchronous generation and predictable demand \cite{panciatici2014advanced}. However, the rapid growth of inverter-based resources, variable renewables, and flexible demand from distributed energy resources (DER) and electric vehicles introduces stochasticity and fast dynamics. Voltage stability, reactive power support, and congestion management; largely ignored in DC approximations, have become increasingly critical for secure and reliable operations \cite{yang2018fundamental}. This shift underscores the need for new approaches that can yield AC-feasible OPF solutions with tracable computational burden.

Data-driven methods have emerged as a promising complement or alternative to classical optimization \cite{wang2025comprehensive}. With the growing availability of PMU measurements \cite{phadke2017phasor} and operational data, machine learning (ML) models have been used to accelerate or approximate AC OPF. Broadly, ML approaches fall into two categories \cite{khaloie2025review}:
(1) \textit{End-to-end (E2E)} learning \cite{baker2022emulating}, which maps operating conditions directly to OPF setpoints, bypassing the solver entirely. E2E approaches have shown strong performance in near-optimal OPF prediction using various architectures, including feedforward networks \cite{baker2022emulating}, graph neural networks \cite{falconer2022leveraging,liu2022topology}, and physics-informed models \cite{huang2022applications,nellikkath2022physics}. However, they often suffer from feasibility violations, limited interpretability, and large data requirements \cite{abelezele2025empirical}.
(2) \textit{Learning-to-optimize (L2O)} \cite{chen2022learning,sun2019survey} on the other hand, utilizes ML to accelerate optimization by predicting active constraint sets, learning surrogate models, or providing warm starts for solvers \cite{baker2019learning} reducing solver iterations and improving feasibility.

In parallel, a distinct body of research on feasibility restoration has emerged, focusing on using ML methods to project the solutions generated from relaxations or approximations into AC solutions. This research direction reinforces the
concept that high-quality initial points or warm start, rather
than stand alone surrogate end-to-end solvers, hold the most
potential \cite{hasan2020survey}. While DC-OPF is widely used in operations as a fast baseline, using it as a systematic physics-informed initialization for AC OPF remains underexplored \cite{huang2022applications}. Existing work has focused mainly on heuristic corrections, such as adding synthetic loads to account for losses \cite{garcia2019general} or tuning susceptance parameters to partially recover voltage effects \cite{taheri2024improving}. Although these methods can bring solutions closer to AC feasibility, they lack robustness and generalizability \cite{khaloie2025review,babiker2025optimal}.

Building on existing machine learning paradigms, this paper proposes a hybrid framework that uses ML models as residual correction models for mapping DC-OPF solutions to their AC OPF counterparts. Our methodology aims to utilize two well established concepts: DC OPF solvers that are widely adopted and trusted by the community and the ability of residual learning to correct for systematic model errors, which has been strongly and repeatedly demonstrated across other fields \cite{cao2023residual, he2016deep, tatsuoka2025deep}. The method first computes a solution using a conventional DC OPF solver. This solution, along with system features, then serves as input for a deep learning model that predicts the final AC OPF solution. This positions our work as a distinct learning paradigm that bypasses the need for an AC OPF solver while being grounded in a physics-based approximation by using the DC OPF solver. Framing the problem as a residual learning task offers better stabilization for the training process, improves generalization across unseen topologies or instances, and reduces data requirements by focusing the model on correcting the systematic errors of a well understood baseline.

The remainder of this paper is organized as follows: Section II introduces the AC and DC OPF formulations and the residual learning concept. Section III describes the proposed methodology, Section IV details the implementation and case studies, and Section V presents the results.
\section{AC and DC Optimal Power Flow Formulations}
\label{sec:opf_formulations_residual}
This section presents the mathematical formulation of the AC and DC OPF problems and how they are formulated as a residual learning task.
\subsection{Notation and Network Model}
Let $\mathcal{B}$ denote the set of buses and $\mathcal{L}\subseteq\mathcal{B}\times\mathcal{B}$ the set of transmission lines. 
For each $b\in\mathcal{B}$, the bus voltage magnitude and angle are denoted by $u_b$ and $\theta_b$, the generator active and reactive powers by $p_{g,b}$ and $q_{g,b}$, and the net active and reactive demands by $p_{d,b}$ and $q_{d,b}$. 
Each branch $(b,n)\in\mathcal{L}$ is characterized by series conductance $g_{bn}$, susceptance $b_{bn}$, and MVA limit $s^{\max}_{bn}$. 
All decision variables are collected as
\begin{equation}
x \equiv \big(p_g,\,q_g,\,u,\,\theta\big)\in{R}^{d},
\end{equation}
\noindent where $d$ is the total number of decision variables determined by the number of buses, generators, and network branches; the exogenous parameters describing loads, topology, and generator costs are denoted by $z$.
\subsection{AC OPF}
\label{subsec:acopf}
The AC OPF problem attempts to determine the operating point that minimizes total generation cost while satisfying power balance equations, voltage limits, and line flow constraints. With a separate convex generation cost $f_{\text{cost}}(p_g)=\sum_{b\in\mathcal{B}} f_b(p_{g,b})$, it is formulated as \cite{grainger1999power, cain2012history}: 

\begin{equation}
\label{eq:acopf_obj_block}
\begin{aligned}
\min_{x}\quad 
& \sum_{b\in\mathcal{B}} f_b(p_{g,b}) \\[1ex]
\end{aligned}
\end{equation}
\begin{equation}
\begin{aligned}
\text{s.t.}\quad 
& p_{g,b}-p_{d,b} = \sum_{n:(b,n)\in\mathcal{L}} p_{bn}(u,\theta),
&& \forall b\in\mathcal{B}, \\[0.5ex]
& q_{g,b}-q_{d,b} = \sum_{n:(b,n)\in\mathcal{L}} q_{bn}(u,\theta),
&& \forall b\in\mathcal{B}, \\[0.5ex]
& \underline{u}_b \le u_b \le \overline{u}_b, \quad 
  \underline{\theta}_{bn}\le \theta_b-\theta_n \le \overline{\theta}_{bn},
&& \forall b\in\mathcal{B}, \\[0.5ex]
& \underline{p}_{g,b}\le p_{g,b}\le \overline{p}_{g,b},\quad 
  \underline{q}_{g,b}\le q_{g,b}\le \overline{q}_{g,b},
&& \forall b\in\mathcal{B}, \\[0.5ex]
& \sqrt{p_{bn}(u,\theta)^2 + q_{bn}(u,\theta)^2}\ \le\ s^{\max}_{bn},
&& \forall (b,n)\in\mathcal{L}.
\end{aligned}
\end{equation}

where active and reactive branch flows follow

\begin{equation}
\label{eq:acopf_pflow}
\begin{split}
p_{bn}(u,\theta)
&= u_b^2 g_{bn}
 - u_b u_n\bigl[g_{bn}\cos(\theta_b-\theta_n) \\
&\qquad\qquad\quad +\, b_{bn}\sin(\theta_b-\theta_n)\bigr]
\end{split}
\end{equation}

\begin{equation}
\label{eq:acopf_qflow}
\begin{split}
q_{bn}(u,\theta)
&= -u_b^2 b_{bn}
 - u_b u_n\bigl[g_{bn}\sin(\theta_b-\theta_n) \\
&\qquad\qquad\quad -\, b_{bn}\cos(\theta_b-\theta_n)\bigr]
\end{split}
\end{equation}

The above constraints enforce nodal active and reactive power balances; 
in addition to voltage magnitude, angle difference, and generator capability limits; and ensure that apparent power branch flows remain within their thermal ratings. 

\subsection{DC OPF}
\label{subsec:dcopf}
The DC OPF is a linear approximation of the AC power flow equations under standard assumptions: unit voltage magnitudes, small phase-angle differences in close nodes, and neglecting line losses and reactive power. 
The resulting active-power balances are \cite{grainger1999power}:
\begin{equation}
p = B\,\theta,\qquad p = p_g - p_d,
\label{eq:dcopf_balance}
\end{equation}
where $B$ is the bus susceptance matrix. 
The corresponding optimization problem is:
\begin{align}
\min_{p_g,\theta}\quad 
& \sum_{b\in\mathcal{B}} f_b(p_{g,b})
\label{eq:dcopf_obj}\\[0.2em]
\text{s.t.}\quad 
& p_g - p_d = B\,\theta, 
\label{eq:dcopf_eq}\\[-0.3em]
& \underline{p}_{g,b}\le p_{g,b}\le \overline{p}_{g,b}, \nonumber\\[-0.3em]
& \big|\tfrac{1}{x_{bn}}(\theta_b-\theta_n)\big|\le f^{\max}_{bn},\quad \forall (b,n)\in\mathcal{L}.
\label{eq:dcopf_ineq}
\end{align}

Let $x_{\text{DC}}(z)\equiv\big(p_g^{\text{DC}}(z),\,\theta^{\text{DC}}(z)\big)$ denote the corresponding DC OPF optimal point.

\vspace{-0.2cm}

\subsection{Residual Learning Between AC and DC OPF}
\label{subsec:residuals}
The DC OPF can be interpreted as a linear approximation of the AC OPF. It captures active power transfer through phase angles but neglects voltage magnitude variations, reactive power balance, and network losses. Consequently, its solutions are generally infeasible under AC constraints. The difference between the AC and DC optimal solutions quantifies these nonlinear effects.

To express this difference, we initialize an approximate AC operating point from the DC OPF solution as:
\begin{equation}
x^{(0)}_{\text{AC}}(z) = \big(p_g^{\text{DC}},\, q_g^{(0)},\, u^{(0)},\, \theta^{\text{DC}}\big),
\label{eq:init_ac_from_dc}
\end{equation}
where $u^{(0)}=\mathbf{1}$ and $q_g^{(0)}=\mathbf{0}$ denote nominal voltage magnitudes and reactive powers. 
The residual between the AC and DC solutions is then defined as
\begin{equation}
\Delta x^{\star}(z) = x^{\star}_{\text{AC}}(z) - x^{(0)}_{\text{AC}}(z).
\label{eq:residual_def}
\end{equation}
This residual $\Delta x^{\star}(z)$ represents the systematic correction required to transform the DC OPF approximation into an AC-feasible operating point.

\subsection{Residual Learning Formulation}
\label{subsec:residual_learning}
Given historical OPF instances, a data-driven model can be trained to approximate the mapping $z \mapsto \Delta x^{\star}(z)$, referred to as the residual learner. The predicted AC solution is then reconstructed as:
\begin{equation}
\widehat{x}_{\text{AC}}(z)=x^{(0)}_{\text{AC}}(z)+\hat{\Delta x}(z),
\label{eq:residual_reconstruct}
\end{equation}
where $\hat{\Delta x}(z)$ denotes the learned residual correction.  
This formulation focuses the model’s representational capacity on capturing the nonlinear deviations ignored by the DC approximation, rather than learning the full AC mapping from scratch. By grounding predictions in a physics-based baseline, the residual learner enhances stability, interpretability, and generalization, while yielding AC-feasible solutions \cite{cao2023residual}.

\section{Methodology}
\label{sec:methodology}

\begin{figure*}[t]
    \centering
    \includegraphics[width=0.8\linewidth]{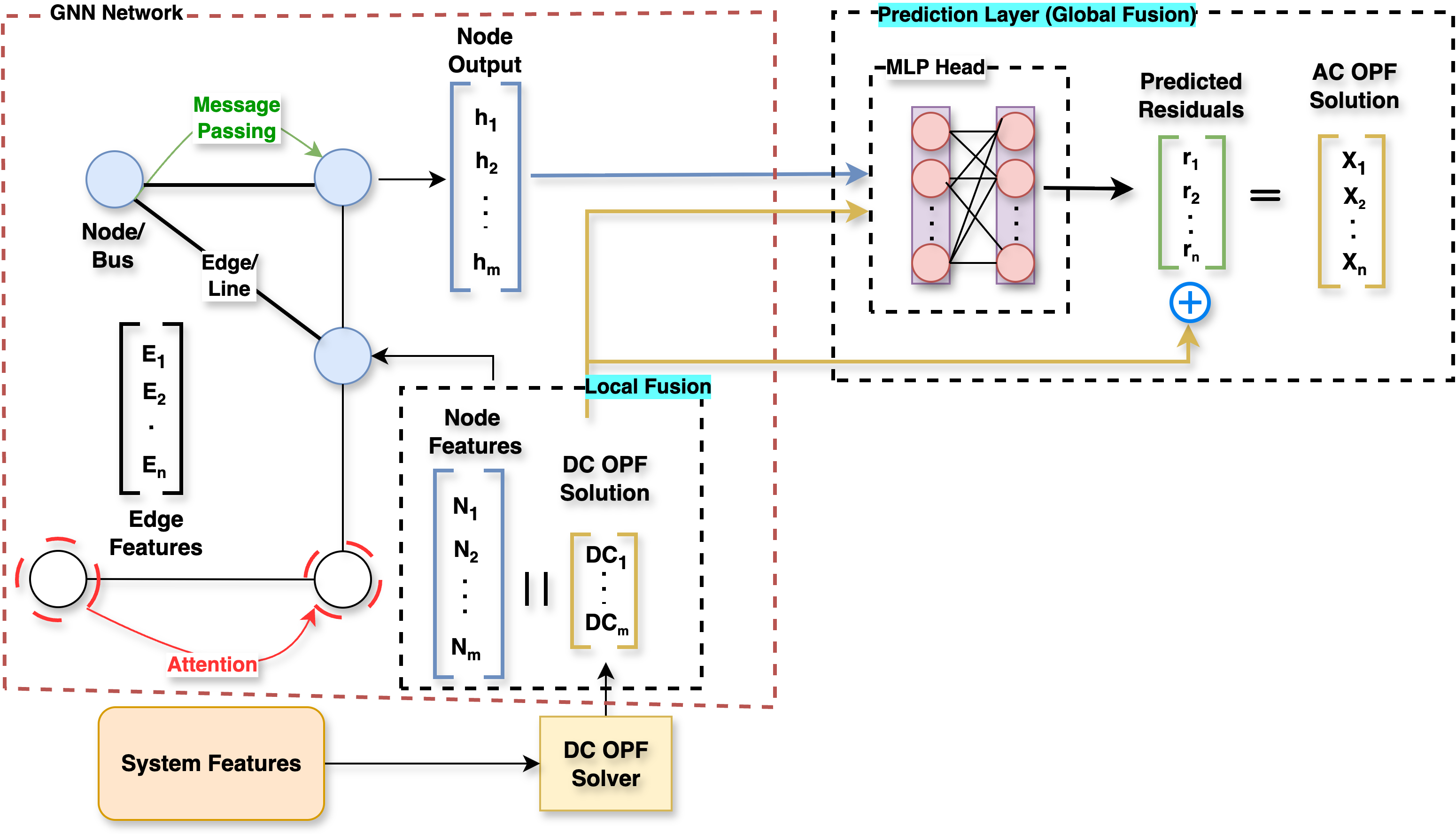}
    \caption{Pipeline for the proposed methodology}
    \label{fig:meth}
\end{figure*}

This section details the proposed residual‐learning framework that maps DC OPF solutions to AC‐feasible operating points. The methodology integrates DC features within a topology‐aware Graph Neural Network (GNN) and learns residual corrections through a physics‐aware objective. GNNs are particularly suitable here because power systems are naturally represented as graphs, where buses and lines correspond to nodes and edges, and message passing mirrors how electrical information propagates through the network; enabling both physical interpretability and scalability. Figure \ref{fig:meth} illustrates the methodology which will be described in this section.

\subsection{Overview}
The proposed model receives as input a power system represented as a graph \cite{xia2021graph}, along with DC OPF variables, and then predicts the residual corrections needed to obtain the AC OPF solution when added to the DC OPF solution. The framework consists of: (i)~DC feature integration, (ii)~a topology-aware local attention GNN with typed message passing, (iii) residual prediction heads using Multi Percepton Layer (MLP) \cite{lecun2015deep}, and (iv)~a physics‐aware training loss enforcing AC feasibility \cite{piloto2024canos}.

\subsection{DC OPF Feature Integration and Global Fusion}
The DC OPF solution provides a fast, physics‐based approximation that captures active power but neglects voltage and reactive effects. 
To best exploit the information of the DC solution, we compute for each operating point the DC variables
\begin{equation}
y^{\mathrm{DC}}=\{\theta_i^{\mathrm{DC}},\,p_{Gi}^{\mathrm{DC}},\,F_{ij}^{\mathrm{DC}}\},
\end{equation}
and integrate them into the model at two levels.

\paragraph{Local Integration}
At the input stage, DC solutions are concatenated with node and edge features before message passing:
\begin{equation}
d_i^{\mathrm{DC}}=[\theta_i^{\mathrm{DC}},\,p_i^{\mathrm{inj,DC}}],\qquad
d_{ij}^{\mathrm{DC}}=[F_{ij}^{\mathrm{DC}}],
\end{equation}
forming $[x_i\!\Vert\! d_i^{\mathrm{DC}}]$ and $[x_{ij}\!\Vert\! d_{ij}^{\mathrm{DC}}]$, where $\!\Vert\!$ represents concatenation. 
This allows the GNN to process DC-informed features aggregated with its neighbor buses (via message passing), ensuring that local corrections are grounded in physical representations.

\paragraph{Global Fusion}
After message passing, the node embeddings are concatenated again with the full DC OPF solution vector before being fed to the MLP head:
\begin{equation}
[h_i^{(K)}\!\Vert\!y^{\mathrm{DC}}] \;\longrightarrow\; \text{MLP head}.
\end{equation}
This step reintroduces the DC operating state globally at the prediction layer, giving the final prediction network access to the baseline operating point while combining it with learned graph embeddings to predict the residuals.  
The two level integration of the DC OPF solution, first in message passing and then in the prediction head, reinforces the connection between the local topology and the global system context.  
Skip connections \cite{he2016deep} are also employed throughout to stabilize training and to emphasize residual corrections relative to the DC baseline solution.

\subsection{Topology‐Aware Local Attention GNN with Typed Messages}
We employ a bus-focused GNN architecture where the adjacency structure directly matches the electrical connectivity of the power network, helping in scaling to multiple topology configurations and enhancing the physical interpretability.

\subsubsection{Attention-Based Message Passing}
We adopt a GNN whose message-passing process mirrors the physical connectivity of the grid. 
The adjacency structure follows the electrical network, allowing each bus to exchange information only with its directly connected neighbors, therefore enforcing locality and physical interpretability.
The model employs an \emph{attention-based message-passing} mechanism, where each neighboring node contributes to the target node's update with a learned weight that reflects its relative electrical importance.

Node and edge embeddings are first initialized using MLPs:
\begin{equation}
h_i^{(0)}=\phi_n([x_i\!\Vert\!d_i^{\mathrm{DC}}]),\qquad 
e_{ij}^{(0)}=\phi_e([x_{ij}\!\Vert\!d_{ij}^{\mathrm{DC}}]).
\end{equation}
For each message-passing layer $\ell$, attention coefficients are computed as:
\begin{equation}
\tilde{\alpha}_{ij}^{(\ell)}=
\frac{(W_q^{(\ell)}h_i^{(\ell)})^\top
      W_k^{(\ell)}[h_j^{(\ell)}\!\Vert\!e_{ij}^{(\ell)}]}{\sqrt{d_k}}
+\psi_{\text{geo}}(x_{ij})
+\psi_{\text{dc}}(F_{ij}^{\mathrm{DC}}),
\end{equation}
and normalized using a softmax over all neighbors $j\!\in\!\mathcal{N}(i)$:
\begin{equation}
\alpha_{ij}^{(\ell)}=
\mathrm{softmax}_{j\in\mathcal{N}(i)}\big(\tilde{\alpha}_{ij}^{(\ell)}\big).
\end{equation}

Each node then aggregates information from its neighbors according to the learned attention weights:
\begin{equation}
m_i^{(\ell)}=
\sum_{j\in\mathcal{N}(i)}
\alpha_{ij}^{(\ell)}W_v^{(\ell)}[h_j^{(\ell)}\!\Vert\!e_{ij}^{(\ell)}].
\end{equation}

The node embedding is subsequently updated as:
\begin{equation}
h_i^{(\ell+1)}=
\mathrm{LN}\!\big(h_i^{(\ell)}+
\sigma(W_h^{(\ell)}[h_i^{(\ell)}\!\Vert\!m_i^{(\ell)}])\big),
\end{equation}
where $\sigma(\cdot)$ denotes a nonlinear activation (ReLU in this work\cite{lecun2015deep}) and $\mathrm{LN}(\cdot)$ denotes Layer Normalization, which stabilizes training and accelerates convergence.

Typed messages are used to handle heterogeneous edge relations—\texttt{ac\_line}, \texttt{transformer}, \texttt{generator\_link}, and \texttt{load\_link}—each with its own learnable parameters.  
This allows the model to distinguish between physically distinct connections such as electrical power transfer and generator-to-bus coupling.  
By combining attention and type-specific message passing, the network learns how different components (i.e transformers, generators) of the system affect one another while conroming to the physical structure of the grid.

\subsection{Residual Prediction Heads}
After $K$ layers of attention-based message passing, each bus is represented by an embedding $h_i^{(K)}$ that captures its local electrical and topological context.  
However, local message passing alone provides only a neighborhood-limited perspective, whereas AC OPF corrections often depend on global operating conditions, such as the total demand and network congestion.
To incorporate this broader view, we compute a pooled graph embedding:
\begin{equation}
z_{\mathcal{G}}=\mathrm{Pool}\big(\{h_i^{(K)}\}\big),
\end{equation}
which summarizes system-wide information by aggregating all bus embeddings via using mean pooling.

Residuals are then predicted using small multilayer perceptrons (MLPs), denoted by 
$\rho_{\text{bus}}(\cdot)$, $\rho_{\text{gen}}(\cdot)$, and $\rho_{\text{br}}(\cdot)$, each mapping concatenated feature vectors to residual outputs.  
For each bus $i$ and branch $(i,j)$:

\begin{equation}
\begin{aligned}
\Delta u_i &= 
\rho_{\text{bus}}\!\big([h_i^{(K)} \!\Vert\! z_{\mathcal{G}} \!\Vert\! y^{\mathrm{DC}}]\big),
\quad u_i \in \{v_i,\theta_i,q_{Gi}\},\\[0.3em]
\Delta p_{Gi} &= 
\rho_{\text{gen}}\!\big([h_i^{(K)} \!\Vert\! z_{\mathcal{G}} \!\Vert\! y^{\mathrm{DC}}]\big),\\[0.3em]
\Delta w_{ij} &= 
\rho_{\text{br}}\!\big([h_i^{(K)} \!\Vert\! h_j^{(K)} 
\!\Vert\! e_{ij}^{(K)} \!\Vert\! z_{\mathcal{G}} \!\Vert\! y^{\mathrm{DC}}]\big),
\quad w_{ij}\in\{|S_{ij}|\}.
\end{aligned}
\end{equation}

Each $\rho(\cdot)$ represents an MLP with its own parameters, which is composed of  fully connected layers with nonlinear activations and dropout for regularization \cite{lecun2015deep}.  
Concatenating the local embedding $h_i^{(K)}$ with the global summary $z_{\mathcal{G}}$ enables the MLP to jointly reason about localized effects and system-level behaviors (e.g. redispatch patterns).  
The integration of $y^{\mathrm{DC}}$ at this level reintroduces the physics-based baseline, grounding the residual predictions in the DC OPF operating point.
It is worth mentioning that we empirically discovered that one MLP to predict all residuals performed better than an MLP for each residual prediction at each bus.

Finally, AC predictions are expressed as residual corrections applied to the DC initialization:
\begin{equation}
\hat u_i = u_i^{(0)} + \Delta u_i, \qquad
\hat w_{ij} = w_{ij}^{(0)} + \Delta w_{ij}.
\end{equation}

Skip connections between $u_i^{(0)}$ and $\Delta u_i$ enforce the residual formulation and ensure the network learns deviations.

\subsection{Loss Function}
The network learns to predict residual corrections on top of the DC OPF baseline; all AC quantities are expressed as:
\begin{equation}
\hat y = y^{(0)} + \Delta y,
\end{equation}
where $y^{(0)}$ represents the DC OPF solution and $\Delta y$ is the learned residual output from the MLP heads.  
The loss function therefore measures both the accuracy of the residual corrections and their physical consistency with AC power flow.

\begin{equation}
\begin{aligned}
\mathcal{L} &=
\mathcal{L}_{\text{sup}}
+\lambda_{\text{pf}}\mathcal{L}_{\text{PF}}
+\lambda_{\text{box}}\mathcal{L}_{\text{box}}\\
&\quad
+\lambda_{\text{obj}}\mathcal{L}_{\text{obj}}
+\lambda_{\text{res}}\mathcal{L}_{\text{res}},
\end{aligned}
\end{equation}

We define five loss components to guide training: a supervised loss to ensure prediction accuracy, physical constraints to enforce feasibility, operational limits, economic optimality, and regularization.

\textbf{Supervised fit:}
This term ensures the residual predictions $\hat y = y^{(0)} + \Delta y$ closely match the true AC labels, hence enforcing data-level fidelity.
\begin{equation}
\begin{aligned}
\mathcal{L}_{\text{sup}} &=
\sum_i \Bigl(
\alpha_v \|\hat v_i - v_i^\star\|_2^2
+ \alpha_\theta \|\hat \theta_i - \theta_i^\star\|_2^2
+ \alpha_q \|\hat q_{Gi} - q_{Gi}^\star\|_2^2 \\
&\quad
+ \alpha_p \|\hat p_{Gi} - p_{Gi}^\star\|_2^2
\Bigr)
+ \sum_{(i,j)} \alpha_S
\Bigl\|\,|\hat S_{ij}| - |S_{ij}^\star|\,\Bigr\|_2^2 .
\end{aligned}
\end{equation}

\textbf{AC power-flow residuals:}
This penalizes violations of the power flow, ensuring the learned residuals generate physically feasible AC operating points.
\begin{equation}
\begin{aligned}
r_i^P &= 
\hat p_{Gi}-p_{Di}
-\hat v_i\!\sum_{j}\hat v_j
\big(G_{ij}\cos\Delta\hat\theta_{ij}
    +B_{ij}\sin\Delta\hat\theta_{ij}\big),\\[-0.3em]
r_i^Q &= 
\hat q_{Gi}-q_{Di}
-\hat v_i\!\sum_{j}\hat v_j
\big(G_{ij}\sin\Delta\hat\theta_{ij}
    -B_{ij}\cos\Delta\hat\theta_{ij}\big),\\[-0.3em]
\mathcal{L}_{\text{PF}} &= 
\sum_i\!\big[(r_i^P)^2+(r_i^Q)^2\big].
\end{aligned}
\end{equation}

\textbf{Operational limits:}
This term enforces voltage, reactive power, and thermal constraints, enhancing the feasibility.
\begin{equation}
\begin{aligned}
\mathcal{L}_{\text{box}} &=
\sum_i\!\big(
[\,\hat v_i-v_i^{\max}\,]_+^2+[\,v_i^{\min}-\hat v_i\,]_+^2\\[-0.3em]
&\quad+[\,\hat q_{Gi}-q_{Gi}^{\max}\,]_+^2+[\,q_{Gi}^{\min}-\hat q_{Gi}\,]_+^2
\big)
+\sum_{(i,j)}[\,|\hat S_{ij}|-S_{ij}^{\max}\,]_+^2.
\end{aligned}
\end{equation}

\textbf{Economic optimality:}
This equation constrains the residual corrected dispatch to remain close to the true OPF cost optimal solution.
\begin{equation}
\mathcal{L}_{\text{obj}}=
\big|C(\hat p_G)-C(p_G^\star)\big|.
\end{equation}

\textbf{Regularization:}
This penalizes large residual magnitudes and reduces the possibility of overfitting \cite{lecun2015deep}.
\begin{equation}
\mathcal{L}{\text{res}}=
\sum_i|\Delta u_i|2^2
+\sum{(i,j)}|\Delta w{ij}|_2^2.
\end{equation}

\section{Case Studies and Experimental Evaluation}
\label{sec:case_studies}

\subsection{Dataset (OPFData)}
In this work we use OPFData, the large-scale AC OPF dataset introduced by Google DeepMind ~\cite{lovett2024opfdata}. OPFData is built from a set of PGLib base networks and, for each selected topology, provides on the order of $3\times10^5$ perturbed operating points with solved AC OPF labels (computed via PowerModels.jl/IPOPT \cite{coffrin2018powermodels, wachter2006implementation}). Each sample is a JSON formatted output that encodes the grid (parameters and topology) and the corresponding AC solution. The graph schema is heterogeneous: nodes \{\texttt{bus}, \texttt{generator}, \texttt{load}, \texttt{shunt}\} and edges \{\texttt{ac\_line}, \texttt{transformer}, \texttt{generator\_link}, \texttt{load\_link}, \texttt{shunt\_link}\}. Labels are provided for buses ($v,\theta$), generators ($p_g,q_g$), and branches (flows on \texttt{ac\_line}/\texttt{transformer}). The dataset include FullTop and N-1 per variants where they enforce a loss of a transmission line or generator.

For training, we adopt the common split of (e.g., $270$k / $15$k / $15$k train/val/test per topology). Experiments will cover the IEEE~57 test system, IEEE~118 test system, and a large-scale (``2000-bus'') case, with additional $N\!-\!1$ line-outage variants for scalability evaluation.

For comprehensive evaluation, we conduct the following case studies:
\begin{itemize}
    \item \textbf{Residual Learning Performance on the IEEE~118 bus system:}
We first evaluate the proposed DC to AC residual learning framework on the IEEE~118-bus network.  
Three variants of graph-based models are compared: a standard GCN (proposed in \cite{owerko2020optimal}) - referred to as Model A - a Graph Attention Network (GAT) \cite{zhang2025gat} - referred to as Model B - and the proposed Local-Attention GNN with residual learning.  
For each architecture, experiments are conducted both with and without residual learning, using identical data splits and hyperparameters.

For the method outlined in Section III, the DC solutions are incorporated at both levels, but the final MLP head predicts the full solution. For the other two methods, DC OPF solutions are incorporated at the final layer when evaluated with residual learning.

\item \textbf{Scalability under $N\!-\!1$ Topological Variations:}
We test model robustness by evaluating the trained model on $N\!-\!1$ contingency variants of the IEEE~118 system, in which transmission lines are removed or generators are unavailable for dispatch. We use the N-1 variations dataset provided in \cite{lovett2024opfdata}. 

\item \textbf{Data Generation using Trained Residual Model:}
Once the model is trained, the residual learner can generate approximate AC OPF solutions directly from DC OPF inputs and system parameters, without additional AC solves.  
As shown in figure \ref{fig:data}, to generate a new training instance a single DC OPF problem is solved to obtain $(p_G^{\text{DC}},\theta^{\text{DC}},F^{\text{DC}})$, which are then passed through the trained model to predict the full AC solution.  

\begin{figure}
    \centering
    \includegraphics[width=0.8\linewidth, height=5cm]{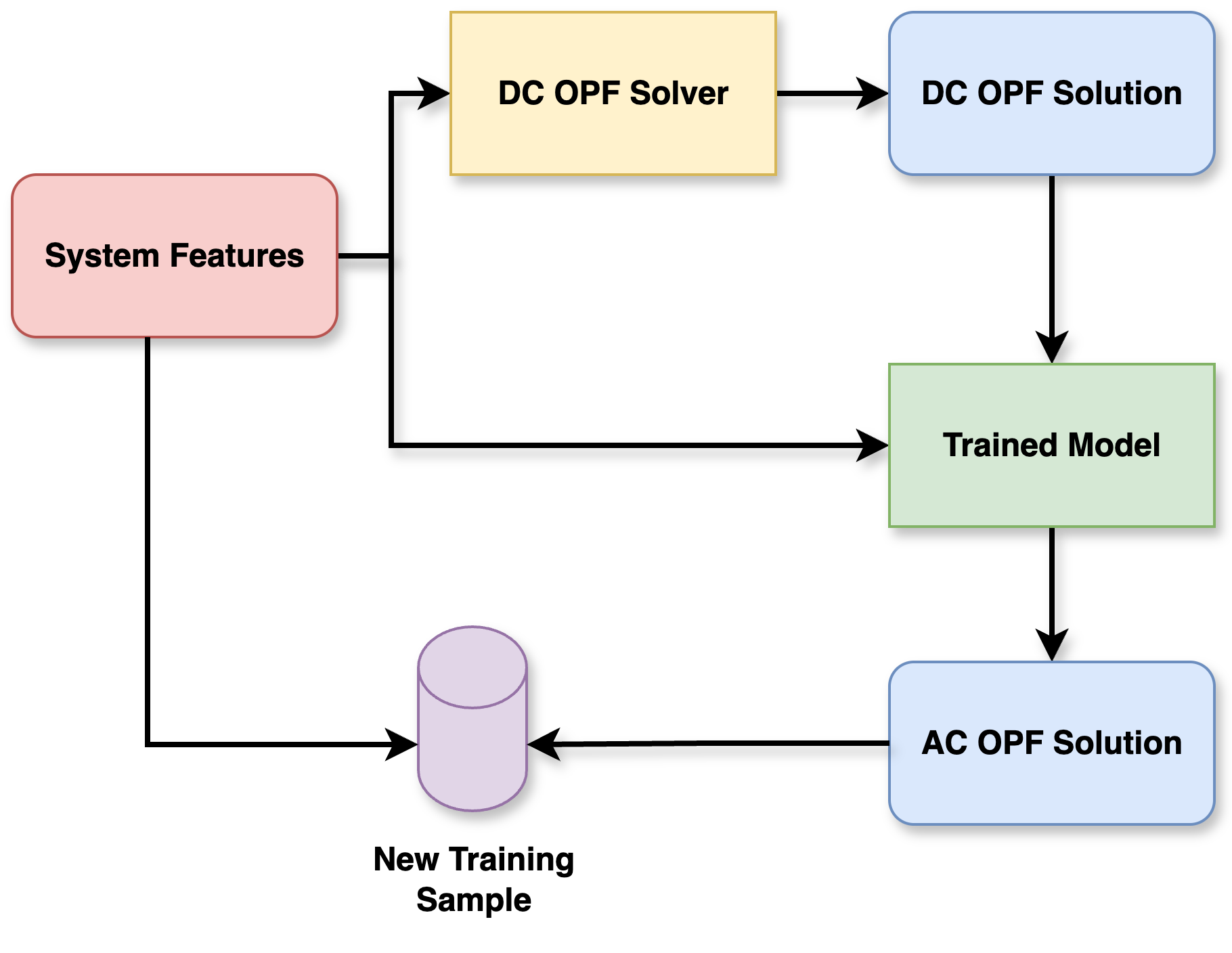}
    \caption{Pipeline for new data generation}
    \label{fig:data}
\end{figure}

\item \textbf{Overall Comparison Across Systems:}
We further benchmark the proposed method against the two baselines on the IEEE~57, IEEE~118, and GOC (Grid Optimization Competition)~2000 bus systems \cite{holzer2024grid} to evaluate scalability on bigger systems.


\end{itemize}

\subsection{Implementation Details}
All DC OPF are solved using the PyPower python package with the IPOPT nonlinear solver \cite{zimmerman2010matpower}.  
The graph-based models are implemented in PyTorch\cite{imambi2021pytorch}, with attention layers and typed message passing implemented via PyTorch Geometric.  
Training is conducted on NVIDIA A100 GPUs with Adam optimization\cite{adam2014method}, early stopping, and gradient clipping for stability\cite{he2016deep}.

Finally, the models performances are assessed using Mean Squared Error (MSE) and feasibility distance which is computed as the average $L_2$ norm of the active and reactive power-flow residuals across all buses. The code to implement the proposed algorithms and the benchmarks on OPFdata is made available for public on this link \footnote{https://github.com/HodgeLab/AC-OPF-residual-learning}.

\section{Results and Discussion}
\label{sec:results}

This section presents the experimental results for the proposed DC to AC residual learning framework across multiple benchmark systems.  
Each subsection corresponds to one of the case studies described in Section~\ref{sec:case_studies}.

\subsection{Residual Learning on IEEE~118}
First, we evaluate the effectiveness of residual learning as outlined Section IV.B on the IEEE 118 bus system.
Table \ref{tab:residual} compares the performance with and without residual learning. It is worth mentioning the proposed model without residuals was implemented as described in section III, but to predict the full solution, not the residuals (the final MLP heads predicts the full solution), however the DC-OPf solutions are integrated in the features of the nodes.

\begin{table}[h]
\label{tab:residual}
\caption{MSE on the IEEE 118 bus system with and without residual learning approach}
\centering
\begin{tabular}{ccccc}
\hline
\textbf{Case}                                                                                & \textbf{Model}                                            & \textbf{\begin{tabular}[c]{@{}c@{}}Bus \\ Voltage\end{tabular}} & \textbf{\begin{tabular}[c]{@{}c@{}}Bus \\ Power\end{tabular}} & \textbf{Feasibility} \\ \hline
\multirow{3}{*}{\textbf{\begin{tabular}[c]{@{}c@{}}Without Residual\\ Learnig\end{tabular}}} & Model A                                                   & 5.45 e-4                                                        & 4.36 e-4                                                      & 4.17 e-4             \\
                                                                                             & Model B                                                   & 5.12 e-4                                                        & 4.53 e-4                                                      & 4.59 e-4             \\
                                                                                             & \begin{tabular}[c]{@{}c@{}}Proposed \\ Model\end{tabular} & 4.72e-4                                                         & 4.11 e-4                                                      & 3.86 e-4             \\ \hline
\multirow{3}{*}{\textbf{\begin{tabular}[c]{@{}c@{}}With Residual\\ Learning\end{tabular}}}   & Model A                                                   & 3.92 e-4                                                        & 3.21 e-4                                                      & 2.52 e-4             \\
                                                                                             & Model B                                                   & 3.88 e-4                                                        & 3.34 e-4                                                      & 2.56 e-4             \\
                                                                                             & \begin{tabular}[c]{@{}c@{}}Proposed \\ Model\end{tabular} & 3.12 e-4                                                        & 2.71e-4                                                       & 1.93 e-4             \\ \hline
\end{tabular}
\end{table}

The results in Table~\ref{tab:residual} demonstrate the effect of incorporating residual learning into all three model variants. 
Without residual learning, the baseline GCN (Model A) and GAT (Model B) exhibit MSE values on the order of $5\times10^{-4}$, while the proposed local-attention GNN slightly improves on these errors even in its non-residual form. 
When residual learning is integrated, the error across all evaluation categories, i.e. bus voltage, bus power, and feasibility, drops by roughly \textbf{35--45\%}, with the proposed model achieving the lowest overall MSE ($3.1\times10^{-4}$ in voltage, $2.7\times10^{-4}$ in power). 
This confirms that anchoring the network to a physics-based DC baseline (DC OPF solutions) allows the training to focus the algorithms representational power on the nonlinear residuals regardless of the underlying architecture.

\subsection{Scalability Over $N\!-\!1$ Topological Variations}
In this section we evaluate the models robustness on N-1 variations on the IEEE 118-bus system. We first evaluate the models without any additional training on the N-1 variations and only trained on the full topology variation. Then we train the model on samples from the N-1 variations.
Table~\ref{tab:robustness} summarizes performance across sampled $N\!-\!1$ cases with and without additional training data.

\begin{table}[h]
\label{tab:robustness}
\caption{Performance on the IEEE 118 system on N-1 variations with and without training}
\centering
\begin{tabular}{ccccc}
\hline
\textbf{Case}                                                                                   & \textbf{Model}                                            & \textbf{\begin{tabular}[c]{@{}c@{}}Bus \\ Voltage\end{tabular}} & \textbf{\begin{tabular}[c]{@{}c@{}}Bus \\ Power\end{tabular}} & \textbf{Feasibility} \\ \hline
\multirow{3}{*}{\textbf{\begin{tabular}[c]{@{}c@{}}Without Additional\\ Training\end{tabular}}} & Model A                                                   & 1.31 e-3                                                        & 1.45 e-3                                                      & 0.97 e-2             \\
                                                                                                & Model B                                                   & 3.62 e-3                                                        & 3.79 e-3                                                      & 0.84 e-2             \\
                                                                                                & \begin{tabular}[c]{@{}c@{}}Proposed \\ Model\end{tabular} & 9.27 e-4                                                        & 1.06e-3                                                       & 2.31 e-3             \\ \hline
\multirow{3}{*}{\textbf{\begin{tabular}[c]{@{}c@{}}With Additional\\ Training\end{tabular}}}    & Model A                                                   & 5.78 e-4                                                        & 4.26 e-4                                                      & 3.82 e-4             \\
                                                                                                & Model B                                                   & 5.11 e-4                                                        & 4.39 e-4                                                      & 4.03 e-4             \\
                                                                                                & \begin{tabular}[c]{@{}c@{}}Proposed \\ Model\end{tabular} & 3.87 e-4                                                        & 3.413 e-4                                                     & 3.23 e-4             \\ \hline
\end{tabular}
\end{table}

Table~\ref{tab:robustness} evaluates how the models generalize to contingency cases when one line or generator is removed. 
When evaluated without any additional fine-tuning, both baseline models experience increase in error, whereas the proposed residual GNN maintains feasibility distances below $2.5\times10^{-3}$ emphasizing the importance of warm starts using DC OPF solutions.
After fine-tuning on $N{-}1$ dataset for the same system (full data), the residual model achieves nearly the same accuracy as in the base topology, with less than 5\% degradation. 
After fine-tuning, the 3 models showed noticeable improvement in performance.

\subsection{Data Generation from DC OPF Solutions}
\begin{figure}[h]
    \centering
    \includegraphics[width=0.95\linewidth]{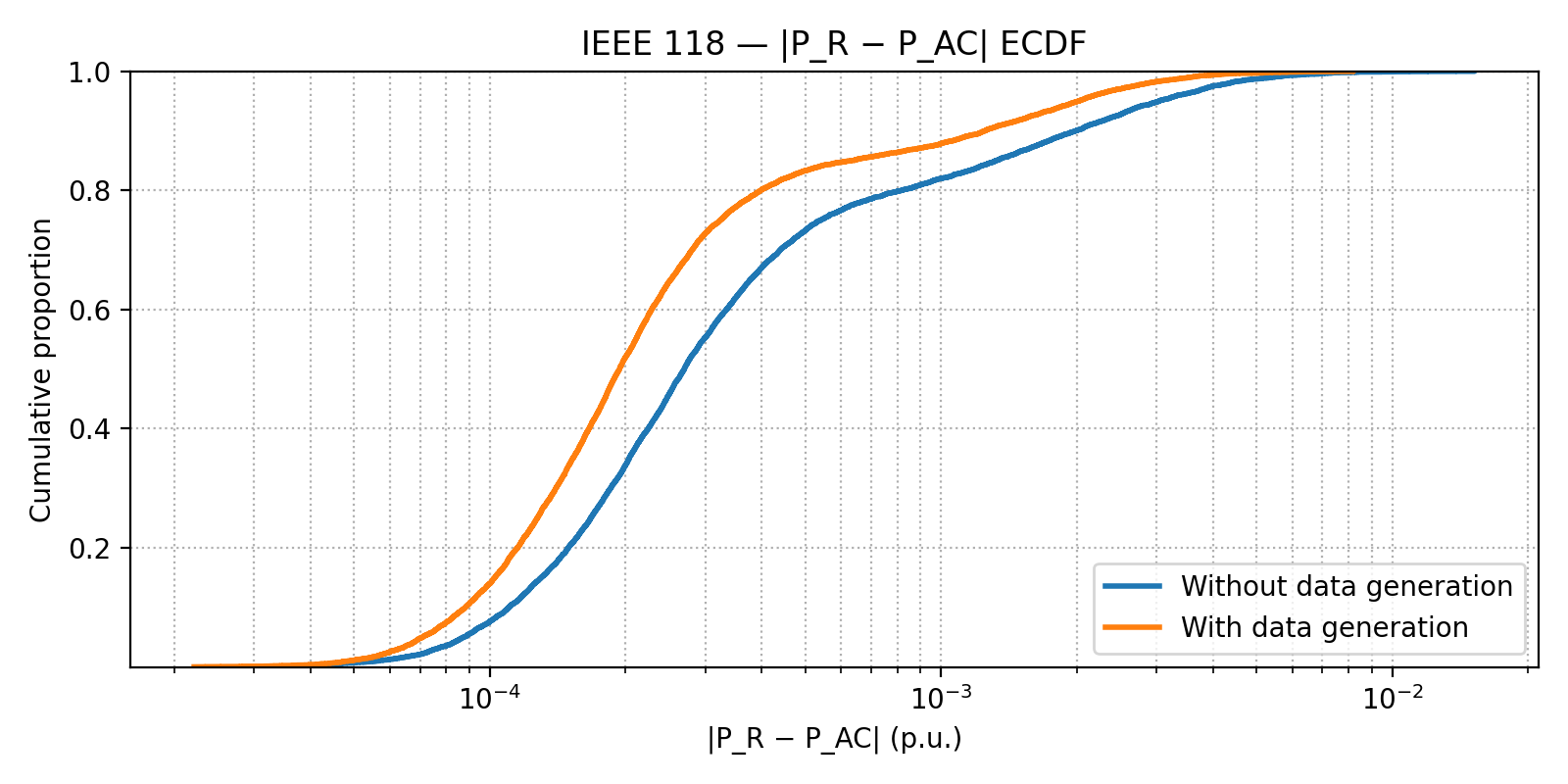}
    \caption{Cumulative proportion of the absolute error in the power predicted with and without additional training data}
    \label{fig:ecdf_power}
\end{figure}

\begin{figure}[h]
    \centering
    \includegraphics[width=0.95\linewidth]{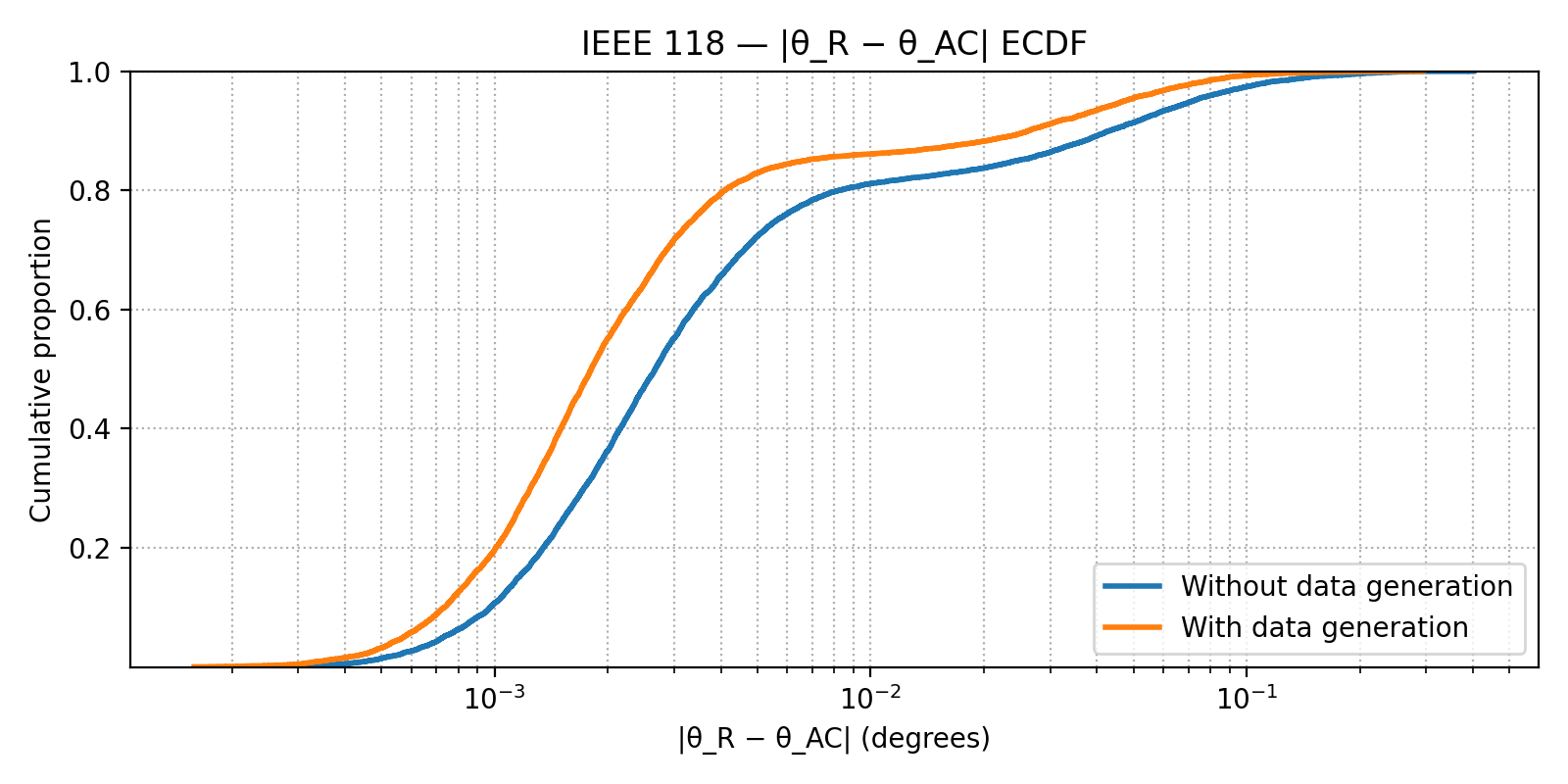}
    \caption{Cumulative proportion of the absolute error in the voltage angle predicted with and without additional training data}
    \label{fig:ecdf_voltage_angle}
\end{figure}

In this section, we use the trained model as a data generator, creating AC-feasible approximations directly from DC OPF inputs to generate new training samples and evaluate if they enhance the overall performance on the IEEE 118-bus system. 
Figures \ref{fig:ecdf_power} and \ref{fig:ecdf_voltage_angle} shows the distribution of the error

Figures~\ref{fig:ecdf_power} and \ref{fig:ecdf_voltage_angle} show the cumulative distribution of absolute prediction errors for generated samples, comparing training with and without additional synthetic data from the residual correction model. 
The curves shift noticeably leftward when the generated data are included, meaning a higher proportion of samples achieve low prediction error. 
Specifically, more than 80\% of voltage-angle predictions and 85\% of power predictions fall within $10^{-3}$ p.u.\ of the AC-OPF labels. 
This verifies that the trained residual learner can serve as a high-quality data generator for AC OPF.

\subsection{Scalability evaluation}
This case study evaluates the computational scalability of the models, evaluated on three systems; IEEE 57-bus system, IEEE 118-bus system and the GOC 2000-bus system.

\begin{table}[h]
\label{tab:overall}
\caption{MSE of the models on the IEEE 57-bus , IEEE 118-bus and GOC 2000-bus systems}
\centering
\begin{tabular}{ccccc}
\hline
\textbf{Case}                          & \textbf{Model}                                            & \textbf{Bus Voltage} & \textbf{Bus Power} & \textbf{Feasibility} \\ \hline
\multirow{3}{*}{\textbf{IEEE 57 Bus}}  & Model A                                                   & 4.43 e-4             & 4.12 e-4           & 3.57 e-4             \\
                                       & Model B                                                   & 4.92 e-4             & 4.14 e-4           & 4.18 e-4             \\
                                       & \begin{tabular}[c]{@{}c@{}}Proposed \\ Model\end{tabular} & 2.62 e-4             & 2.71e-4            & 1.93 e-4             \\ \hline
\multirow{3}{*}{\textbf{IEEE 118 Bus}} & Model A                                                   & 5.45 e-4             & 4.36 e-4           & 4.17 e-4             \\
                                       & Model B                                                   & 5.12 e-4             & 4.53 e-4           & 4.59 e-4             \\
                                       & \begin{tabular}[c]{@{}c@{}}Proposed \\ Model\end{tabular} & 3.12 e-4             & 2.71e-4            & 1.93 e-4             \\ \hline
\multirow{3}{*}{\textbf{GOC 2000 Bus}} & Model A                                                   & 2.34 e-3             & 3.45 e-3           & 2.91 e-3             \\
                                       & Model B                                                   & 2.5 e-3              & 2.41 e-3           & 2.03 e-3             \\
                                       & \begin{tabular}[c]{@{}c@{}}Proposed \\ Model\end{tabular} & 1.04 e-3             & 1.31 e-3           & 0.949 e-4            \\ \hline
\end{tabular}
\end{table}

Table~\ref{tab:overall} compares all three models across the IEEE 57-bus, 118-bus, and 2000-bus systems.
Error magnitudes grow slightly with network size as expected due to higher dimensionality, but the residual GNN consistently achieves the lowest MSE and feasibility errors. 
On the largest 2000-bus system, it reduces voltage and power MSEs by nearly 50\% compared with Model A and Model B, demonstrating strong scalability and numerical stability.

\subsection{Run Time Analysis}

This section presents the run times for AC OPF, DC OPF and the proposed method. An NVIDIA A100 GPU was used to generate the predictions for the proposed method, whereas a server with 128 CPUs and 64 GB of RAM was used to run the AC and DC OPF using the PyPower \cite{zimmerman2010matpower} Python package.

\begin{table}[h]
\label{tab:runtime}
\caption{Solve time of AC-IPOPT, DC-IPOPT and proposed method on the test cases}
\centering
\begin{tabular}{ccc}
\hline
\textbf{Case}                          & \textbf{Model} & \textbf{Solve Time (ms)} \\ \hline
\multirow{3}{*}{\textbf{IEEE 57 Bus}}  & DC-IPOPT       &           24               \\
                                       & AC-IPOPT       &            337              \\
                                       & Proposed Model &            138              \\ \hline
\multirow{3}{*}{\textbf{IEEE 118 Bus}} & DC-IPOPT       &            113              \\
                                       & AC-IPOPT       &            2467              \\
                                       & Proposed Model &            231             \\ \hline
\multirow{3}{*}{\textbf{GOC 2000 Bus}} & DC-IPOPT       &           6345               \\
                                       & AC-IPOPT       &           92856               \\
                                       & Proposed Model &            7134             \\ \hline
\end{tabular}
\end{table}

Table~\ref{tab:runtime} compares computational performance between conventional IPOPT-based DC and AC solvers and the proposed model inference on a GPU. 
The residual GNN offers one to two orders of magnitude acceleration relative to AC OPF.
For instance, on the IEEE 118 system, AC OPF solution with IPOPT requires about 2.47 s, whereas the residual GNN inference takes 0.23~s, achieving nearly 11$\times$ speedup. 
The benefit becomes even more evident on the 2000-bus system, where AC OPF solution with IPOPT takes almost 93~s compared to 7~s for the proposed model. 
These results confirm that machine learning models have great potential for accelerating AC OPF calculation in real time decision making.

\section{Conclusion}
\label{sec:conclusion}

This work presented a residual learning framework for mapping DC OPF solutions to their AC-feasible counterparts using a topology-aware, attention-based Graph Neural Network. 
By embedding DC solutions within both the message-passing and prediction stages and enforcing physics-aware constraints during training, the proposed method captures nonlinear corrections neglected by DC approximations while preserving physical interpretability. 
Experiments on the OPFData benchmark demonstrate that the framework improves accuracy, AC feasibility, and scalability across the IEEE 57-bus, IEEE 118-bus, and GOC 2000-bus systems, as well as being robust to $N{-}1$ topological variations. 
The approach also significantly accelerates AC-feasible solution generation and enables direct use of DC OPF outputs for high-fidelity approximations to generate new training samples. 
These findings position residual learning as a promising bridge between traditional optimization and modern data-driven methods for real time power system operations. 

Future research will extend this framework toward probabilistic, multi-period OPF and time-series based OPF, in addition to the development of foundation models pre-trained across large-scale power-system datasets.

\bibliographystyle{ieeetr}
\bibliography{sources}

\thanksto{\noindent Submitted to the 24th Power Systems Computation Conference (PSCC 2026).}
\end{document}